\documentclass[]{interact}

\usepackage{epstopdf}% To incorporate .eps illustrations using PDFLaTeX, etc.
\usepackage[caption=false]{subfig}% Support for small, `sub' figures and tables

\usepackage{amsmath,amssymb,amsfonts}
\usepackage{graphicx}
\usepackage{textcomp}
\usepackage{xcolor}
\usepackage{tikz}
\usetikzlibrary{
    shapes.geometric, % For rectangles, ellipses
    arrows.meta,      % For advanced arrow tips
    positioning,      % For 'right=of', 'below=of'
    calc              % For coordinate calculations
}

\usepackage[numbers,sort&compress]{natbib}% Citation support using natbib.sty
\bibpunct[, ]{[}{]}{,}{n}{,}{,}% Citation support using natbib.sty
\renewcommand\bibfont{\fontsize{10}{12}\selectfont}% Bibliography support using natbib.sty
\makeatletter% @ becomes a letter
\def\NAT@def@citea{\def\@citea{\NAT@separator}}% Suppress spaces between citations using natbib.sty
\makeatother% @ becomes a symbol again

\tikzset{
    crop_img/.style={draw, thick, inner sep=1pt, line cap=round},
    encoder/.style={draw, thick, rectangle, fill=blue!10, minimum height=3.5cm, minimum width=2.5cm, text width=2.2cm, align=center, font=\bfseries},
    head/.style={draw, thick, rectangle, fill=gray!20, minimum height=1cm, minimum width=2.5cm, text width=2.2cm, align=center},
    loss/.style={draw, thick, ellipse, fill=red!10, minimum width=2cm, align=center},
    update_box/.style={draw, thick, rectangle, fill=green!10, font=\bfseries},
    main_arrow/.style={-{[round]Stealth[length=8pt, width=10pt]}, thick},
    ema_arrow/.style={-{[round]Stealth[length=8pt, width=10pt]}, thick, dashed},
    grad_arrow/.style={-{[round]Stealth[length=8pt, width=10pt]}, thick, red}
}

\begin{document}

\articletype{RESEARCH ARTICLE}

\title{MedDChest: A Content-Aware Multimodal Foundational Vision Model for Thoracic Imaging}

\author{
\name{Mahmoud Soliman\textsuperscript{a}\thanks{Mahmoud Soliman. Email: mosama97@student.ubc.ca}, Islam Osman\textsuperscript{a}, Mohamed S. Shehata\textsuperscript{a} and Rasika Rajapakshe\textsuperscript{a,b,c}}
\affil{\textsuperscript{a}Department of Computer Science, Mathematics, Physics and
Statistics, The University of British Columbia,  
Kelowna, BC, Canada; \textsuperscript{b}Medical Physics, BC Cancer, Kelowna, BC, Canada; \textsuperscript{c}Department of Surgery, The University of British Columbia, Vancouver, BC, Canada}
}

\maketitle

% %%% Input sections from the 'sec' directory %%%
% Note: sec/0_abstract.tex should contain the abstract and keywords environments.
\begin{abstract}
The performance of vision models in medical imaging is often hindered by the prevailing paradigm of fine-tuning backbones pre-trained on out-of-domain natural images. To address this fundamental domain gap, we propose MedDChest, a new foundational Vision Transformer (ViT) model optimized specifically for thoracic imaging. We pre-trained MedDChest from scratch on a massive, curated, multimodal dataset of over 1.2 million images, encompassing different modalities including Chest X-ray and Computed Tomography (CT) compiled from 10 public sources. A core technical contribution of our work is Guided Random Resized Crops, a novel content-aware data augmentation strategy that biases sampling towards anatomically relevant regions, overcoming the inefficiency of standard cropping techniques on medical scans. We validate our model’s effectiveness by fine-tuning it on a diverse set of downstream diagnostic tasks. Comprehensive experiments empirically demonstrate that MedDChest significantly outperforms strong, publicly available ImageNet-pretrained models. By establishing the superiority of large-scale, in-domain pre-training combined with domain-specific data augmentation, MedDChest provides a powerful and robust feature extractor that serves as a significantly better starting point for a wide array of thoracic diagnostic tasks. The model weights will be made publicly available to foster future research and applications.
\end{abstract}

\section{Introduction}
\label{sec:introduction}

Deep learning has become a cornerstone of modern medical image analysis, with Vision Transformers (ViTs)~\cite{Dosovitskiy20_ViT} pushing the boundaries of what is possible by capturing long-range dependencies within images~\cite{Shen17}. The dominant paradigm involves fine-tuning models pre-trained on large-scale, out-of-domain datasets like ImageNet~\cite{Deng09}. However, this approach is fundamentally limited by the significant domain gap between natural and medical images. Natural images are typically 3-channel (RGB) photographs, whereas thoracic scans are often single-channel grayscale representations of complex anatomy. This discrepancy in feature distribution and color space forces the model to expend significant capacity to adapt, limiting its ultimate performance on downstream medical tasks. We posit that pre-training from scratch on a massive, in-domain dataset is a more effective strategy for building truly foundational models for medicine.

To address this challenge, we introduce \textbf{MedDChest}, a new foundational ViT model specifically optimized for thoracic imaging. Our central thesis is that large-scale, self-supervised pre-training on in-domain data provides a superior feature representation. To build MedDChest, we aggregated a massive, multimodal dataset of over 1.2 million images from 10 public sources, encompassing Chest X-rays and CT scans. This large-scale effort, similar in spirit to recent work~\cite{Gupta24_MedMAE}, allows MedDChest to learn a rich visual vocabulary directly relevant to thoracic anatomy. Our most significant technical contribution is a novel data augmentation strategy, Guided Random Resized Crop, designed to solve a critical flaw in applying standard self-supervised learning (SSL) to medical scans. Methods like Masked Autoencoders (MAE)~\cite{He22_MAE} are inefficient on medical images, which often contain large, non-informative backgrounds. Our two-stage algorithm first identifies the content's bounding box and then biases the crop sampling process towards this region, dramatically improving the efficiency and quality of pre-training.

This paper introduces MedDChest, a powerful foundational model for thoracic imaging, and details the large-scale dataset curated for its pre-training. We present our novel content-aware augmentation technique, Guided Random Resized Crop, and provide rigorous empirical validation of its effectiveness. By demonstrating significant performance gains over strong ImageNet-pretrained baselines, we establish MedDChest as a superior starting point for downstream tasks, and we will release our model weights to facilitate future research.
\section{Related Work}
\label{sec:related_work}

Our research builds upon established work in transfer learning, self-supervised learning (SSL), and the application of Vision Transformers (ViTs) to medicine. While fine-tuning ImageNet-pretrained models like ResNet~\cite{He16_ResNet} and DenseNet~\cite{Huang17_DenseNet} has been highly successful on medical tasks~\cite{Gulshan16_Retinopathy, Rajpurkar17_CheXNet}, the domain gap between natural and medical images remains a known limitation~\cite{Raghu19_Transfusion, Azizi21_RevisitingTransfer}. This has motivated a shift towards creating foundational models pre-trained directly on in-domain medical data, primarily through SSL.

SSL allows models to learn from vast unlabeled datasets, which is critical in medicine. While early contrastive methods like SimCLR~\cite{Chen20_SimCLR} were effective, recent work has favored masked image modeling approaches like Masked Autoencoders (MAE)~\cite{He22_MAE}, especially for ViTs~\cite{Dosovitskiy20_ViT}. The success of this paradigm has spurred large-scale medical pre-training initiatives. For instance, RadImageNet~\cite{Mei22_RadImageNet} aggregated millions of radiology images for pre-training CNNs, while MedMAE~\cite{Gupta24_MedMAE} proposed a generalist backbone for diverse medical tasks, both validating the benefits of in-domain pre-training.

Our work is distinct from these efforts in two crucial ways. First, we focus specifically on the thoracic region to develop a more potent, specialized feature set. Second, and most importantly, we identify and solve a key methodological flaw in applying SSL to medical scans. Standard augmentation like Random Resized Crop is inefficient for medical images, which often contain large empty backgrounds. While some have used computationally heavy saliency maps to guide cropping, our work introduces an efficient, algorithm-driven solution, \textbf{Guided Random Resized Crop}. By performing a rapid content analysis before sampling, we ensure the model consistently learns from relevant anatomy, thereby maximizing the effectiveness of the pre-training process.
\section{Methodology}
\label{sec:methodology}
% Add this to your document preamble:
% \usepackage{tikz}
% \usepackage{graphicx}
% \usetikzlibrary{shapes.geometric, arrows.meta, positioning, fit, backgrounds}

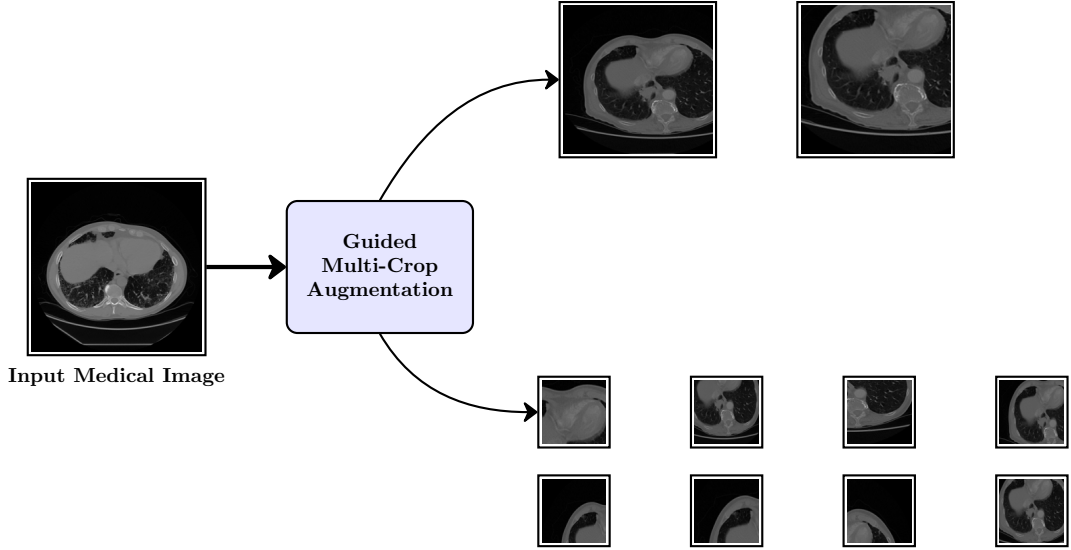
\begin{figure}[htbp]
\centering
% Declare the background layer
\pgfdeclarelayer{background}
\pgfsetlayers{background,main}

% Simply add the 'scale' option here. All other dimensions can remain the same.
\begin{tikzpicture}[scale=0.7, transform shape]
    % --- STYLES --- (Can be kept outside the figure for consistency)
    \tikzset{
        node distance=0.5cm and 1.5cm,
        image_frame/.style={
            draw, thick, inner sep=2pt, line cap=round
        },
        process_box/.style={
            draw, thick, rectangle, rounded corners,
            fill=blue!10, minimum height=2.5cm, minimum width=3.5cm,
            text width=3cm, align=center, font=\bfseries
        },
        label_box/.style={
            rectangle, rounded corners, fill=gray!15, inner sep=0.4cm
        },
        main_arrow/.style={
            -{[round]Stealth[length=8pt,width=12pt]},
            ultra thick
        },
        branch_arrow/.style={
            -{[round]Stealth[length=6pt,width=9pt]},
            thick
        }
    }
    % == 1. PLACE THE INPUT IMAGE ==
    % This node holds the full input CT scan
    \node[image_frame] (input_image) {\includegraphics[width=3.2cm]{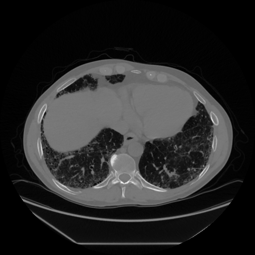}}; % Reduced from 4cm
    \node[below=0.1cm of input_image, font=\bfseries] {Input Medical Image};

    % == 2. PLACE THE AUGMENTATION BOX ==
    % This box represents the process that generates the crops
    \node[process_box, right=of input_image] (augmentation) {Guided Multi-Crop\\Augmentation};

    % == 3. PLACE THE GLOBAL CROPS ==
    % These are the two larger, higher-resolution views
    \node[image_frame, above right=0.8cm and 1.6cm of augmentation] (global1) {\includegraphics[width=2.8cm]{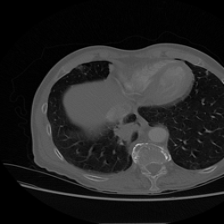}}; % Reduced from 3.5cm
    \node[image_frame, right=of global1] (global2) {\includegraphics[width=2.8cm]{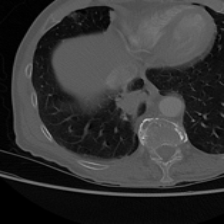}}; % Reduced from 3.5cm

    % == 4. PLACE THE LOCAL CROPS ==
    % These are the eight smaller, lower-resolution views, arranged in a grid
    \node[image_frame, below right=0.8cm and 1.2cm of augmentation] (local1) {\includegraphics[width=1.2cm]{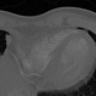}}; % Reduced from 1.5cm
    \node[image_frame, right=of local1] (local2) {\includegraphics[width=1.2cm]{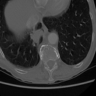}};
    \node[image_frame, right=of local2] (local3) {\includegraphics[width=1.2cm]{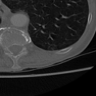}};
    \node[image_frame, right=of local3] (local4) {\includegraphics[width=1.2cm]{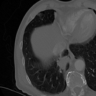}};
    
    \node[image_frame, below=of local1] (local5) {\includegraphics[width=1.2cm]{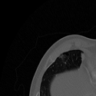}};
    \node[image_frame, right=of local5] (local6) {\includegraphics[width=1.2cm]{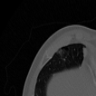}};
    \node[image_frame, right=of local6] (local7) {\includegraphics[width=1.2cm]{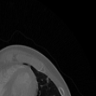}};
    \node[image_frame, right=of local7] (local8) {\includegraphics[width=1.2cm]{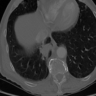}};

    % == 5. DRAW ARROWS TO SHOW THE FLOW ==
    % A thick arrow from the input to the process box
    \draw[main_arrow] (input_image.east) -- (augmentation.west);
    
    % Thinner arrows branching out to the crop groups
    \draw[branch_arrow] (augmentation.north) to[out=60, in=180] (global1.west);
    \draw[branch_arrow] (augmentation.south) to[out=-60, in=180] (local1.west);

    % == 6. DRAW LABELS FOR THE CROP GROUPS (on the background layer) ==
    % This ensures the labels are drawn behind the images for a clean look

\end{tikzpicture}
\caption{Multi-crop augmentation process for medical image analysis. The input medical image is processed to generate multiple views: 2 global crops at higher resolution and 8 local crops at lower resolution for comprehensive feature extraction.}
\label{fig:multi_crop_augmentation}
\end{figure}

\begin{figure*}[htbp]
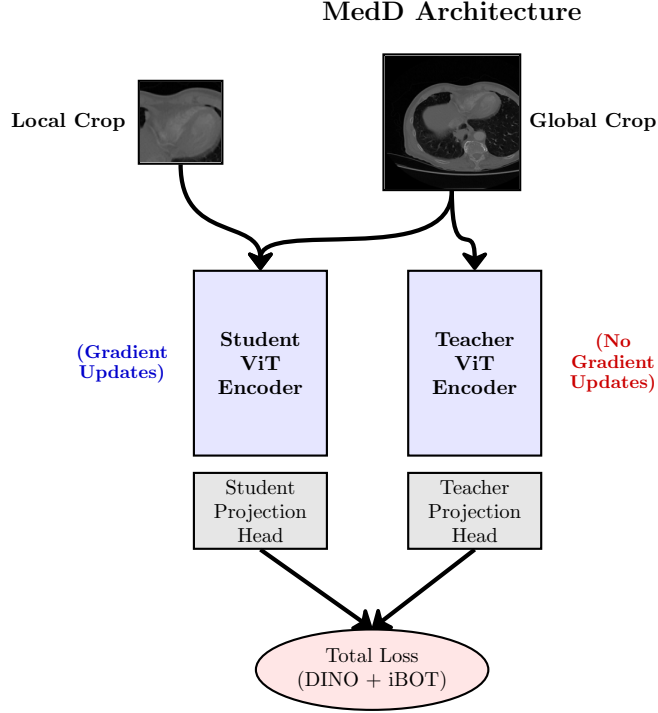

\centering

% Define a set of styles for consistent and clean diagram elements
\tikzset{
    % Style for the input crop image nodes
    crop_img/.style={
        draw, thick, inner sep=1pt, line cap=round
    },
    % Style for the main encoder boxes (Student/Teacher)
    encoder/.style={
        draw, thick, rectangle, fill=blue!10, 
        minimum height=3.5cm, minimum width=2.5cm, 
        text width=2.2cm, align=center, font=\bfseries
    },
    % Style for the smaller projection head boxes
    head/.style={
        draw, thick, rectangle, fill=gray!20,
        minimum height=1cm, minimum width=2.5cm,
        text width=2.2cm, align=center
    },
    % Style for the loss calculation ellipses
    loss/.style={
        draw, thick, ellipse, fill=red!10, 
        minimum width=3cm, minimum height=1.5cm, align=center
    },
    % Style for the final backpropagation box
    update_box/.style={
        draw, thick, rectangle, fill=green!10,
        font=\bfseries, minimum width=2.5cm, minimum height=1cm
    },
    % Style for a standard solid arrow
    main_arrow/.style={
        -{[round]Stealth[length=8pt, width=10pt]}, 
        thick
    },
    % Style for the dashed EMA update arrow
    ema_arrow/.style={
        -{[round]Stealth[length=8pt, width=10pt]}, 
        thick, dashed, blue
    },
    % Style for the red gradient update arrow
    grad_arrow/.style={
        -{[round]Stealth[length=8pt, width=10pt]}, 
        thick, red
    }
}

% Begin the TikZ picture environment
\begin{tikzpicture}[scale=0.7, transform shape]
    % --- STYLES --- (Can be kept outside the figure for consistency)
    \tikzset{
        node distance=0.5cm and 1.5cm,
        image_frame/.style={
            draw, thick, inner sep=2pt, line cap=round
        },
        process_box/.style={
            draw, thick, rectangle, rounded corners,
            fill=blue!10, minimum height=2.5cm, minimum width=3.5cm,
            text width=3cm, align=center, font=\bfseries
        },
        label_box/.style={
            rectangle, rounded corners, fill=gray!15, inner sep=0.4cm
        },
        main_arrow/.style={
            -{[round]Stealth[length=8pt,width=12pt]},
            ultra thick
        },
        branch_arrow/.style={
            -{[round]Stealth[length=6pt,width=9pt]},
            thick
        }
    }
    % == 1. DEFINE INPUT CROPS ==
    \node[crop_img] (local_crop) {\includegraphics[width=1.5cm]{fig/local_crop_1.png}};
    \node[crop_img, right=3cm of local_crop] (global_crop) {\includegraphics[width=2.5cm]{fig/global_crop_1.png}};
    
    % Add labels below the crops
    \node[left=0.1cm of local_crop] {\textbf{Local Crop}};
    \node[right=0.02cm of global_crop] {\textbf{Global Crop}};
    
    % == 2. DEFINE STUDENT AND TEACHER TOWERS ==
    \node[encoder, below=2cm of local_crop, xshift=1.5cm] (student) {Student\\ViT\\Encoder};
    \node[encoder, right=of student] (teacher) {Teacher\\ViT\\Encoder};
    
    % Add the projection heads below the encoders
    \node[head, below=0.3cm of student] (student_head) {Student\\Projection\\Head};
    \node[head, below=0.3cm of teacher] (teacher_head) {Teacher\\Projection\\Head};
    
    % No gradient label for teacher
    \node[font=\small\bfseries, right=0.3cm of teacher, text width=1.8cm, align=center, color=red!80!black] (no_grad_label) {(No Gradient\\Updates)};

    % No gradient label for teacher
    \node[font=\small\bfseries, left=0.3cm of student, text width=1.8cm, align=center, color=blue!80!black] (no_grad_label) {(Gradient\\Updates)};

    % == 3. DRAW ASYMMETRIC INPUT FLOW ==
    % Local crop goes ONLY to the student
    \draw[main_arrow] (local_crop.south) .. controls +(south:1.5cm) and +(north:1cm) .. (student.north);
    
    % Global crop goes to BOTH student and teacher
    \draw[main_arrow] (global_crop.south) .. controls +(south:1.5cm) and +(north:1cm) .. (student.north);
    \draw[main_arrow] (global_crop.south) .. controls +(south:1.5cm) and +(north:1cm) .. (teacher.north);

    % == 4. LOSS CALCULATION ==
    \node[loss, below=1.5cm of student_head, xshift=2.1cm] (total_loss) {Total Loss\\(DINO + iBOT)};

    % Connect the projection heads to the loss function
    \draw[main_arrow] (student_head.south) -- (total_loss.north);
    \draw[main_arrow] (teacher_head.south) -- (total_loss.north);

    % Add method labels
    \node[above=0.5cm of global_crop, font=\Large\bfseries] {MedD Architecture};

\end{tikzpicture}

\caption{DINOv2 Self-Supervised Learning Architecture in MedD. The asymmetric data augmentation strategy feeds global crops to both student and teacher networks, while local crops are only fed to the student network. The teacher network is updated via exponential moving average (EMA) and receives no gradient updates.}
\label{fig:dinov2_architecture_final}
\end{figure*}

Our methodology is designed to build a powerful foundational model, MedDChest, through a meticulously adapted self-supervised learning pipeline tailored to the unique challenges of medical imaging. This section provides a comprehensive account of our approach. We first outline the advanced self-supervised framework that serves as the basis of our work. We then introduce our primary technical contribution, a domain-specific data augmentation strategy named Content-Guided Multi-Crop Augmentation. Finally, we describe the large-scale dataset curated for pre-training and our implementation and evaluation protocols.

\subsection{Self-Supervised Pre-training Framework}
\label{ssec:pretraining_framework} We select the DINOv2~\cite{Oquab23_DINOv2} framework as the foundation for our pre-training regimen, chosen for its demonstrated ability to learn robust, general-purpose visual features without supervision. The framework's architecture, illustrated in Figure~\ref{fig:dinov2_architecture_final}, integrates a student-teacher learning paradigm with a compound loss function derived from the principles of both DINO~\cite{Caron21_DINO} and iBOT~\cite{Zhou21_IBOT}.

\subsubsection{Student-Teacher Architecture}
The learning process relies on a student network, $g_{\theta_s}$, and a teacher network, $g_{\theta_t}$, which share an identical Vision Transformer (ViT)~\cite{Dosovitskiy20_ViT} architecture. The student's parameters, $\theta_s$, are learned through standard backpropagation of the loss function. Conversely, the teacher network is not updated by gradients. Its parameters, $\theta_t$, are instead updated as an exponential moving average (EMA) of the student's parameters. This "mean teacher" approach~\cite{Tarvainen17_MeanTeacher} ensures the teacher provides a more stable learning target, a critical component for preventing training collapse in self-distillation frameworks. The update rule is given by:
\begin{equation}
    \theta_t \leftarrow \lambda \theta_t + (1 - \lambda) \theta_s,
\end{equation}
where $\lambda$ is a momentum coefficient that follows a cosine schedule from 0.996 to 1.0 throughout training. As depicted in Figure~\ref{fig:dinov2_architecture_final}, this ensures that gradients flow exclusively to the student network.

\subsubsection{Compound Loss Function}
The student network is optimized using a compound loss function $\mathcal{L}_{\text{total}}$ that synergistically combines multiple self-supervised objectives. Both the student and teacher networks terminate in a projection head which outputs logits, subsequently normalized into a probability distribution over $K$ dimensions. Let $P_s(x)$ and $P_t(x)$ denote the output distributions for a given image view $x$.

 The primary objective is an image-level cross-entropy loss that encourages representational consistency across different augmented views of an image. For a pair of views $x_i$ and $x_j$, the student is trained to predict the teacher's output distribution. This is formulated as:
\begin{equation}
    \mathcal{L}_{\text{image}} = - \sum_{\substack{(x_i, x_j) \\ i \neq j}} P_t(x_j) \log P_s(x_i).
\end{equation}
To avoid trivial solutions, the teacher's distribution is "sharpened" using a low temperature parameter, $\tau_t$, in its softmax calculation ($P_t(x) = \text{softmax}((g_{\theta_t}(x))/\tau_t)$), while the student uses a higher temperature $\tau_s$. Additionally, a centering operation is applied to the teacher's output before the softmax to prevent dominant features from collapsing the representation~\cite{Caron21_DINO}.

To learn fine-grained local features, we incorporate a patch-level masked image modeling objective inspired by iBOT~\cite{Zhou21_IBOT}. A random subset of the input patches fed to the student network are masked. The student is then tasked to predict the teacher's output features for these masked regions. This is also formulated as a cross-entropy loss, $\mathcal{L}_{\text{patch}}$, between the student's predictions and the teacher's feature distribution for the corresponding unmasked patches.

 To further improve feature quality, DINOv2 introduces a regularization term, the KoLeo loss, which encourages the features across all patches to be utilized uniformly. It minimizes the Kullback-Leibler (KL) divergence between the average feature distribution across all patches in a batch, $\bar{p}$, and a uniform distribution, $\mathcal{U}$:
\begin{equation}
    \mathcal{L}_{\text{KoLeo}} = \text{KL}(\bar{p} || \mathcal{U}).
\end{equation}
The total loss guiding the student's learning is a weighted sum of these three components:
\begin{equation}
    \mathcal{L}_{\text{total}} = \alpha_1 \mathcal{L}_{\text{image}} + \alpha_2 \mathcal{L}_{\text{patch}} + \alpha_3 \mathcal{L}_{\text{KoLeo}}.
\end{equation}

\subsection{Content-Guided Multi-Crop Augmentation}
\label{ssec:guided_cropping}
A central challenge in applying self-supervised learning to medical imaging lies in the standard data augmentation pipeline. The multi-crop strategy~\cite{Caron20_SwAV}, which underpins frameworks like DINOv2, relies on Random Resized Crop. This operation is highly inefficient for medical scans (e.g., CT, X-ray) that often contain large, non-informative background regions or collimation borders. Frequent sampling of these empty areas provides a weak, or even misleading, learning signal, which can hinder the model's ability to learn meaningful anatomical features and slow down convergence.

To address this critical flaw, we introduce a novel two-stage sampling strategy named \textbf{Content-Guided Multi-Crop Augmentation}. This technique, visually detailed in Figure~\ref{fig:multi_crop_augmentation}, ensures that every crop is focused on relevant anatomical content while preserving the randomness essential for effective learning.

\subsubsection{Stage 1: Content Analysis.}
For any given single-channel grayscale input image $I \in \mathbb{R}^{H \times W}$, we first perform a rapid, threshold-based segmentation to identify the region of interest. We derive a binary content mask $M$ by applying a near-zero intensity threshold $\theta$:
\begin{equation}
    M(i, j) = \mathbb{I}(I(i, j) > \theta),
\end{equation}
where $\mathbb{I}(\cdot)$ is the indicator function. This simple operation effectively isolates the foreground anatomical structures from the void background. From this mask, we compute the minimal bounding box $B = (x_{\min}, y_{\min}, x_{\max}, y_{\max})$ that encloses the support of $M$, i.e., all pixels where $M(i,j) = 1$.

\subsubsection{Stage 2: Biased Crop Sampling.}
Standard Random Resized Crop samples the parameters for a crop (area, aspect ratio, and position) relative to the entire image dimensions. Our method modifies this by biasing the positional sampling. First, we define a slightly padded sampling region $B'$ from the content box $B$ to allow for some variability at the edges, ensuring this region is clipped to the original image boundaries. The scale and aspect ratio of the crop are still chosen randomly, following the original DINOv2 protocol. However, the crucial difference lies in the selection of the crop's top-left corner $(i, j)$. Instead of being sampled from a uniform distribution over the entire image plane, $\mathcal{U}([0, H-h] \times [0, W-w])$, it is sampled from a uniform distribution restricted to the valid starting positions within the biased content region $B'$:
\begin{equation}
    (i, j) \sim \mathcal{U}(B').
\end{equation}
This guarantees that every generated crop, whether global or local, is centered on meaningful anatomical structures. This modification dramatically improves the quality and efficiency of the learning signal passed to the student and teacher networks, making the pre-training process far more effective for the medical domain.

\subsection{Pre-training Dataset Curation}
\label{ssec:dataset_curation}
To power our self-supervised pre-training, we aggregated and curated a massive, multi-source dataset comprising over 1.2 million images. A robust foundational model requires exposure to a wide variety of data to learn generalizable features; therefore, our corpus was compiled from 10 major public sources. This ensures a high degree of diversity in imaging modalities (CT, X-ray), anatomical locations (with a focus on the thorax but including others for feature richness), pathologies, and patient demographics. A detailed overview of the constituent datasets is provided in Table~\ref{tab:dataset_overview}.

\begin{table*}[htbp]
    \centering
    \caption{A detailed overview of the various datasets collected to form the pre-training corpus for MedDChest. This table is adapted from prior work~\cite{Gupta24_MedMAE} to ensure methodological comparability and transparency.}
    \label{tab:dataset_overview}
    % \resizebox has been removed to fix the large font issue
    \begin{tabular}{|l|l|r|l|}
        \hline
        \textbf{Collection} & \textbf{Location} & \textbf{Subjects} & \textbf{DataTypes} \\
        \hline
        CheXpert~\cite{Irvin19_CheXpert} & Chest & 65,240 & X-Ray \\
        COVID-19-NY-SBU~\cite{Saltz21_SBU} & Lung & 1,384 & CR, CT, DX \\
        CT images in COVID19~\cite{An20_CovidCT} & Lung & 661 & CT \\
        MIDRC-RICORD~\cite{Tsai21_MIDRC} & Lung & 227 & CT \\
        QIN Lung CT~\cite{Kalpathy15_QIN} & Lung & 47 & CT \\
        LungCT-Diagnosis~\cite{Grove15_LungCT} & Lung & 61 & CT \\
        NSCLC-Radiomics-Genomics~\cite{Aerts14_NSCLC} & Lung & 89 & CT \\
        RIDER Lung CT~\cite{Zhao12_RIDER} & Chest & 32 & CT, CR, DX \\
        LIDC-IDRI~\cite{Armato11_LIDC} & Chest & 1,010 & CT, CR, DX \\
        COVID-19-AR~\cite{Desai20_AR} & Lung & 105 & CT, DX, CR \\
        \hline
    \end{tabular}
\end{table*}

\subsection{Implementation and Evaluation Protocol}
\label{ssec:implementation}

The backbone for our MedDChest model is a standard Vision Transformer (ViT-Base) architecture~\cite{Dosovitskiy20_ViT}, which features 12 transformer layers, 12 self-attention heads, and a 768-dimensional embedding space, using a patch size of 16x16 pixels. A crucial modification was made to the initial patch projection layer, adapting it to natively accept single-channel grayscale inputs, thereby aligning the model's architecture directly with the predominant format of thoracic medical images. The entire pre-training pipeline was implemented in PyTorch~\cite{Paszke19_PyTorch}. Training was conducted on a cluster of 2 NVIDIA A100 GPUs for approximately 14 days, using the AdamW optimizer~\cite{Loshchilov17_AdamW} with a batch size of 128 and a cosine learning rate schedule.

\section{Experiments and Results}
\label{sec:experiments}

To validate MedDChest's effectiveness, we evaluate its learned features on two widely-used public datasets for thoracic imaging and compare our results against state-of-the-art models.

\subsection{Experimental Setup and Evaluation Protocol}
The MedDChest model (\texttt{vit\_base}) was pre-trained from scratch using the DINOv2 methodology, as detailed in Section~\ref{sec:methodology}. Key hyperparameters included a batch size of 64, the AdamW optimizer~\cite{Loshchilov17_AdamW} with a base learning rate of $4 \times 10^{-3}$, and a 1000-epoch training schedule. The pre-training utilized \texttt{fp16} mixed-precision and our novel Guided Random Resized Crop augmentation.

To specifically assess the quality of the learned features, we strictly adhere to the linear probing evaluation protocol~\cite{He22_MAE, Caron21_DINO}. In this setup, the entire pre-trained MedDChest backbone is \textbf{frozen}. Only a single, randomly initialized linear classification layer, trained on top of these static features, is updated. This standard benchmark provides an unbiased estimate of feature representation quality.

We evaluate on two datasets. For the Chest X-ray Pneumonia dataset from Kermany et al.~\cite{Kermany18_Pneumonia}, we report the Area Under the Receiver Operating Characteristic Curve (AUROC). For the NIH ChestX-ray14 dataset~\cite{Wang17_NIH}, we report multi-class accuracy to provide a direct, head-to-head comparison with prior foundational model work, specifically MedMAE~\cite{Gupta24_MedMAE}.

\subsection{Results and Analysis}
On the pneumonia detection task, MedDChest achieves a state-of-the-art AUROC of \textbf{99.8\%} under the linear probing protocol. As shown in Table~\ref{tab:auroc_results}, the ability of a simple linear classifier to achieve such a strong result highlights the exceptional quality of the features learned by MedDChest.

For the NIH ChestX-ray14 task, MedDChest achieves a linear probing accuracy of \textbf{94.5\%} (Table~\ref{tab:accuracy_results}). This result is particularly noteworthy as it surpasses the MedMAE model by a significant margin of \textbf{6.5\%}. Achieving superior performance with a frozen backbone strongly validates our core hypothesis: thoracic-specific pre-training combined with our Guided Random Resized Crop augmentation produces a more powerful and relevant feature representation than a generalist medical model.

% --- TABLE 1: AUROC RESULTS ---
\begin{table}[h]
\centering
\caption{Pneumonia detection performance (AUROC) on the Pneumonia X-ray dataset. Prior works are ordered by performance, with our result presented at the end for comparison.}
\label{tab:auroc_results}
\resizebox{0.8\columnwidth}{!}{%
\begin{tabular}{|l|c|}
\hline
\textbf{Model/Architecture} & \textbf{AUROC (\%)} \\
\hline
Xception~\cite{Stephen19_Xception} & 94.23 \\
Vision Transformer (ViT)~\cite{Singh22_ViT} & 96.00 \\
ResNet34 (Weak Supervision)~\cite{Li21_WeakSupervision} & 99.49 \\
MAE with Adult Pretraining~\cite{Khatami22_MAE} & 99.60 \\
EfficientNetB0~\cite{Hammoudi21_EfficientNet} & 99.70 \\
Weighted Ensemble~\cite{Chouhan20_Ensemble} & 99.76 \\
Lightweight WAE~\cite{Liang20_LightweightWAE} & 99.77 \\
\hline \hline
\textbf{MedDChest (Ours)} & \textbf{99.8} \\
\hline
\end{tabular}%
}
\end{table}

% --- TABLE 2: ACCURACY RESULTS ---
\begin{table}[h]
\centering
\caption{Linear probing accuracy on the ChestX-ray14 dataset. Prior works are ordered by performance}
\label{tab:accuracy_results}
\resizebox{0.7\columnwidth}{!}{%
\begin{tabular}{|l|c|}
\hline
\textbf{Method} & \textbf{Accuracy (\%)} \\
\hline
Wang et al.~\cite{Wang17_NIH} & 73.8 \\
Yao et al.~\cite{Yao17_Scratch} & 79.8 \\
AE-CNN~\cite{Cai20_AECNN} & 82.4 \\
MUXNet-m~\cite{Lu20_MUXNet} & 84.1 \\
LEAF~\cite{Liang21_LEAF} & 84.3 \\
CheXNet~\cite{Rajpurkar17_CheXNet} & 84.4 \\
MedMAE~\cite{Gupta24_MedMAE} & 88.0 \\
\hline \hline
\textbf{MedDChest (Ours)} & \textbf{94.5} \\
\hline
\end{tabular}%
}
\end{table}
\section{Conclusion}
\label{sec:conclusion}

In this work, we introduced \textbf{MedDChest}, a foundational Vision Transformer for thoracic imaging, pre-trained from scratch on over 1.2 million in-domain medical images. Our primary technical contribution, Guided Random Resized Crop, is a novel content-aware augmentation technique designed to overcome the inefficiencies of standard self-supervised learning on medical scans by focusing on relevant anatomical structures.

While this paper focuses on classification, future work will explore MedDChest's utility for segmentation and detection, investigate performance gains from full fine-tuning, and pursue model distillation for clinical deployment. By releasing our model weights, we offer a robust new baseline for thoracic image analysis and a stronger foundation for the next generation of diagnostic AI.

% \section*{Acknowledgements}
% %%% Optional: If you have acknowledgements, add them here or in a separate file. %%%
% The authors would like to thank...

% \section*{Funding}
% %%% Optional: If you have funding information, add it here or in a separate file. %%%
% This work was supported by [Grant Numbers].

% %%% BIBLIOGRAPHY %%%
% This uses the Taylor & Francis NLM bibliography style.
% Make sure your .bib file is named 'main.bib' and that the 'tfnlm.bst'
% style file is available in your project folder.
\bibliographystyle{tfnlm}
\bibliography{main}

\end{document}